\documentclass[conference]{IEEEtran}

\usepackage[T1]{fontenc}
\usepackage{microtype}

\usepackage{amsmath}
\usepackage{amssymb}
\usepackage{amsfonts}

\usepackage{graphicx}
\usepackage{booktabs}
\usepackage{array}
\usepackage{tabularx}
\usepackage{multirow}
\usepackage{makecell}
\usepackage{stfloats}

\usepackage[table]{xcolor}
\usepackage{colortbl}

\usepackage{cite}

\usepackage{hyperref}
\hypersetup{hidelinks}

\begin{document}

\title{
Traffic-MLLM: Curiosity-Regularized Supervised Learning for Traffic Scenario Case-Based Reasoning
}

\author{
Waikit Xiu$^{1,2\dagger}$ \quad
Qiang Lu$^{2\dagger}$ \quad
Bingchen Liu$^{3}$ \quad
Chen Sun$^{1*}$ \quad
Xiying Li$^{2*}$\footnotemark \\[4pt]

$^{1}$The University of Hong Kong, Hong Kong SAR, China \\
$^{2}$Sun Yat-sen University, Shenzhen, China \\
$^{3}$University of Bristol, United Kingdom \\[4pt]

$^{\dagger}$Equal contribution \qquad
$^{*}$Corresponding authors
}

\maketitle

\begin{abstract}
For safe and robust autonomous driving, decision-making systems must effectively leverage past experiences to handle the inherent long-tail of traffic scenarios. Case-Based Reasoning (CBR) provides a natural paradigm for this by adapting solutions from prior cases. However, in complex and dynamic traffic environments, traditional CBR methods struggle to effectively abstract and adapt knowledge under uncertainty. Meanwhile, although multimodal large language models (MLLMs) exhibit strong perceptual and linguistic capabilities, their reasoning behavior often relies on empirical pattern fitting, limiting robustness under distribution shift and long-tail scenarios. We propose \textit{Traffic-MLLM}, a retrieval-free neural case modeling framework for multimodal traffic reasoning. Instead of performing explicit case retrieval at inference time, Traffic-MLLM learns a structured and generalizable case space directly during training. To support this learning process, we construct a multi-source case base by integrating dynamic traffic videos and large-scale static visual question-answering data, serving as a unified training substrate for learning structured case representations. To further improve representation quality near knowledge boundaries, we introduce a curiosity-driven refinement mechanism based on Random Network Distillation (RND), encouraging the model to internalize cross-case structural regularities rather than surface correlations. Experiments on the SUTD-TrafficQA and DriveQA benchmarks demonstrate consistent improvements in dynamic reasoning, regulatory understanding, and cross-domain transfer. Traffic-MLLM achieves 50.8\% accuracy on SUTD-TrafficQA, 74.8\% on the CARLA-based DriveQA split, and 83.1\% on the real-world Mapillary split, indicating that representation-level case-space refinement provides an effective alternative to explicit retrieval for scalable multimodal case adaptation. 

\textit{Code and model weights will be publicly released.}

\end{abstract}

\footnotetext{This work may be submitted to a conference. Copyright may be transferred without notice.}

\section{Introduction}
\label{sec:Introduction}
Case-Based Reasoning (CBR) facilitates the structured representation of prior experiences, modeling similarity relationships among cases and adapting historical precedents to novel contexts to achieve robust generalization~\cite{Aamodt1994,Kolodner1993,Richter2005}. In highly dynamic and complex traffic environments, human driving decisions inherently rely on analogical reasoning over prior experiences, reflecting the underlying understanding and transfer of structure within a case space~\cite{Kolodner1993}. From this perspective, revisiting multimodal traffic reasoning through the lens of CBR provides critical insights into the structural causes of limited generalization observed in contemporary models~\cite{Koh2021WILDS}.

In recent years, Multimodal Large Language Models (MLLMs) have demonstrated significant potential in traffic scene understanding and question-answering tasks~\cite{Alayrac2022Flamingo,Liu2023LLaVA,Sima2024DriveLM}. However, the majority of existing approaches rely on Supervised Fine-Tuning (SFT) or instruction-style tuning, which treats training samples as independent prediction instances rather than organized components within a structured case space~\cite{Ouyang2022InstructGPT}. In traffic scenarios characterized by long-tail distributions and cross-domain variations, such training paradigms often bias learning toward high-frequency statistical patterns and fail to adapt robustly to weakly represented or distribution-shifted scenarios~\cite{Lin2017FocalLoss,Koh2021WILDS}. From a CBR standpoint, this limitation stems not merely from restricted predictive capacity, but from the insufficient organization and utilization of structural information within the case collection~\cite{Aamodt1994}.

To address these challenges, we formalize multimodal traffic data as structured cases and construct a multi-source case base that integrates dynamic video-based cases with static image-based reasoning cases~\cite{Xu2021TrafficQA,Sima2024DriveLM,Dosovitskiy2017CARLA}. Each case encompasses visual context, a textual query, and the corresponding reasoning outcome. Specifically, dynamic cases capture temporal interactions and future-state evolution, while static cases encode regulatory reasoning and fine-grained visual semantics. Crucially, this case base is designed not to support online retrieval or explicit case reuse~\cite{Aamodt1994}, but to serve as a unified training substrate for cross-domain case learning.

Nevertheless, the implementation of a case base does not inherently guarantee effective structural learning. Under standard SFT optimization, models remain predisposed to prioritize high-frequency cases, leading to the underfitting of boundary or weakly represented instances~\cite{Lin2017FocalLoss}. Consequently, the central challenge lies in optimizing the model's capacity to learn from the case base without relying on explicit case retrieval at inference time~\cite{Kolodner1993}.

To this end, we propose a retrieval-free neural case learning paradigm. Rather than performing case matching during inference, we enhance structural learning during training via a curiosity-regularized case-space optimization mechanism~\cite{Burda2018RND}. Specifically, we leverage decoder hidden states as case embeddings and employ Random Network Distillation (RND) to estimate their novelty within the learned manifold, thereby quantifying the relative epistemic position of each case. This novelty signal facilitates the dynamic adjustment of optimization weights, incentivizing the model to allocate greater modeling capacity to boundary and under-represented cases. Through this mechanism, the model transitions from surface-level statistical fitting to the abstraction of cross-case structural regularities, resulting in superior case adaptation under distribution shifts and long-tail conditions~\cite{Koh2021WILDS}.

\begin{figure*}[!tbp] 
    \centering 
    \includegraphics[width=\linewidth]{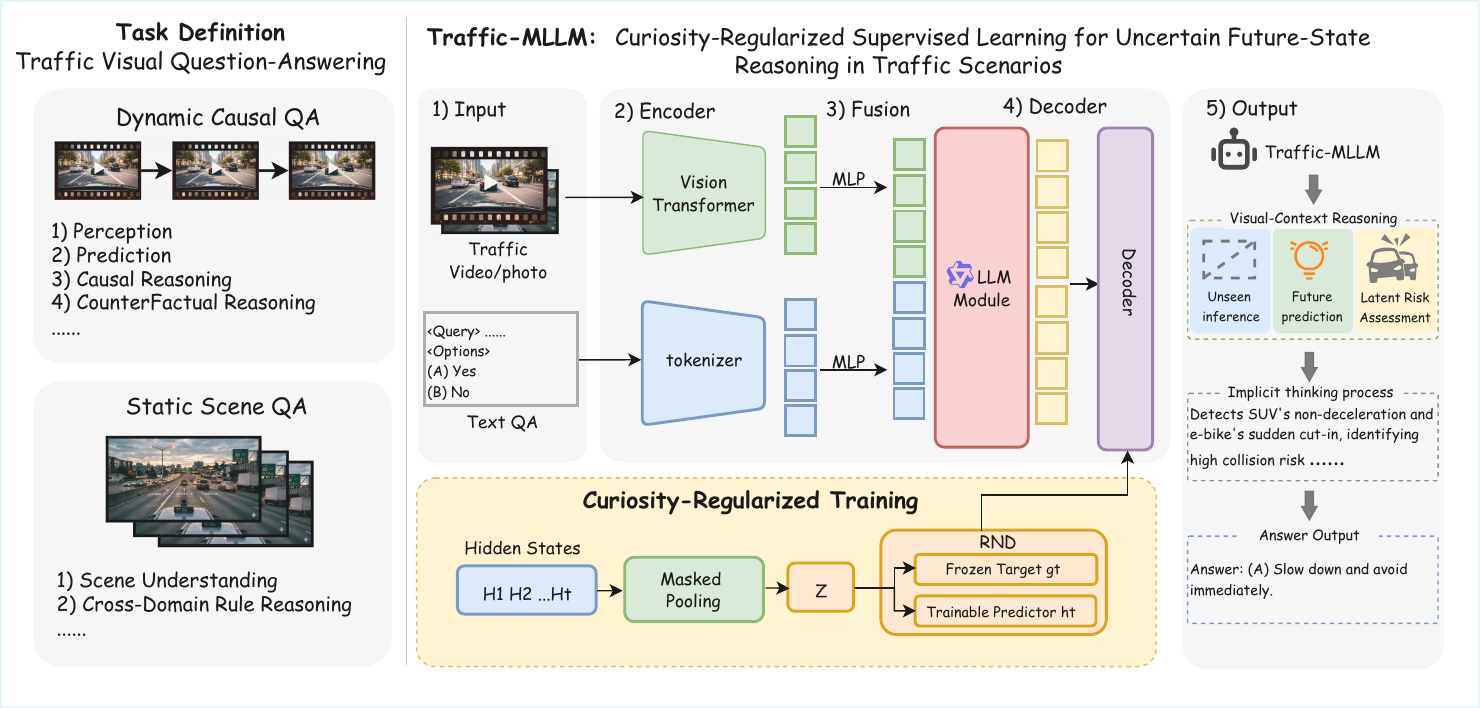} 
    \caption{Overview of Traffic-MLLM. The framework unifies dynamic video reasoning and static image-based question answering through a vision--text encoder--fusion--decoder architecture. Curiosity-regularized training is applied on decoder representations without modifying the forward inference structure.} \
    \label{fig:framework} 
\end{figure*}

\section{Related Work}
\label{sec:related_work}

\subsection{Case-Based Reasoning in Autonomous Systems}
Case-Based Reasoning (CBR) has long been recognized as a paradigm that models problem solving through the reuse and adaptation of prior cases~\cite{Aamodt1994,Kolodner1993}. In autonomous systems, prior studies explored CBR for planning and adaptive decision-making~\cite{Watson1999CBR}. With the emergence of Large Language Models (LLMs), retrieval-augmented reasoning frameworks have been proposed to improve generalization~\cite{Lewis2020RAG}. In driving-oriented tasks, DiLu~\cite{Wen2023DiLu} shows that knowledge-driven memory modules can enhance reasoning. However, representing complex multimodal traffic cases and enabling robust structural adaptation in open-world environments remain challenging.

\subsection{Multi-Source Case Base Construction}
Constructing a structured and diverse case base is critical for enabling cross-domain generalization. Existing traffic datasets span perception, prediction, and reasoning tasks, including TrafficQA~\cite{Xu2021TrafficQA}, DriveLM~\cite{Sima2024DriveLM}, and simulation environments such as CARLA~\cite{Dosovitskiy2017CARLA}. These data sources provide complementary coverage over dynamic interactions, long-tail events, and diverse environmental conditions. Nevertheless, long-tail distributions and cross-domain shifts remain key challenges for robust traffic reasoning systems~\cite{Koh2021WILDS}.

\subsection{Multimodal LLMs for Traffic Understanding}
Recent Multimodal Large Language Models (MLLMs) have significantly advanced visual and multimodal reasoning capabilities~\cite{Liu2023LLaVA}. Extensions to video understanding further enable temporal reasoning over dynamic scenes~\cite{Lin2023VideoLLaVA}. In traffic-specific contexts, multimodal reasoning frameworks have been applied to driving explanation and forecasting~\cite{Sima2024DriveLM}. However, large-scale multimodal models can suffer from shortcut learning and hallucination~\cite{Ji2023Hallucination}. To address the limitations of static supervision, intrinsic motivation mechanisms such as Random Network Distillation (RND) quantify novelty through prediction error~\cite{Burda2018RND}. Building on these ideas, our work introduces curiosity-driven regularization into case-space optimization to improve multimodal traffic reasoning under long-tail and distribution-shift conditions~\cite{Koh2021WILDS}.

\section{Methodology}
\label{sec:methodology}
\subsection{Problem Definition}

We formulate multimodal traffic reasoning as a case-based supervised learning problem. Let $\mathcal{D}=\{C_i\}_{i=1}^{N}$ denote a traffic case base, where each case is defined as
\begin{equation}
    C = (x, q, a, e),
\end{equation}
where $x$ represents the visual context (traffic video or image), $q$ is a natural-language query describing the reasoning task, $a$ is the corresponding answer or decision outcome, and $e$ is an optional explanation. Given an input pair $(x,q)$, the goal of the model is to predict the correct output $a$. Let $f_{\theta}$ denote a parameterized multimodal large language model that models the conditional distribution
\begin{equation}
    p_{\theta}(a \mid x, q).
\end{equation}
Classical case-based reasoning systems typically perform explicit retrieval of similar cases during inference. In contrast, we treat the case base as a structured collection of training experiences used to optimize the representation and decision function of the model. By learning from heterogeneous cases in $\mathcal{D}$, the model progressively organizes multimodal traffic observations within its internal feature space, enabling improved generalization across diverse traffic scenarios, particularly under long-tail events and distribution shifts.

\subsection{Multi-Source Case Base Construction}
Traffic-MLLM reformulates multimodal traffic reasoning as a structured case-space learning problem. Rather than treating training samples as independent supervision instances, we explicitly model each traffic example as a structured case and focus on how such cases are internalized within the model’s representation manifold.

Architecturally, Traffic-MLLM follows a unified vision--text encoder--fusion--decoder pipeline, supporting both dynamic video reasoning and static image-based question answering within a single autoregressive framework. Given an input $x=(v,q)$, where $v\in\mathbb{R}^{T_v\times H\times W\times C}$ denotes a traffic video (or image when $T_v=1$) and $q=(q_1,\dots,q_L)$ denotes a textual query, the visual input is partitioned into spatiotemporal patches and encoded into visual tokens $X_v=\{x_i^v\}_{i=1}^{N_v}$. Each token is associated with a coordinate $(t_i,h_i,w_i)$ to preserve temporal and spatial structure. The textual query is embedded as $X_q=\{x_j^q\}_{j=1}^{L}$.

Visual tokens are projected into the language embedding space via $f_{\text{proj}}(\cdot)$ and concatenated with textual tokens to form a multimodal sequence $X=[X_v',X_q]$, where $X_v'=f_{\text{proj}}(X_v)$. The decoder performs autoregressive generation conditioned on $X$:
\begin{equation}
    p_\theta(y\mid x)=\prod_{t=1}^{T}p_\theta(y_t\mid y_{<t},X).
\end{equation}

To encode spatiotemporal dependencies, each visual token $x_i^v$ is assigned a positional index $p_i=(t_i,h_i,w_i)$ and injected into the decoder’s self-attention through rotary position embeddings:
\begin{equation}
    \alpha_{ij}\propto 
    \exp\left(
    \frac{(Q_i\cdot R(p_i))(K_j\cdot R(p_j))^\top}{\sqrt{d}}
    \right),
\end{equation}
where $Q_i$, $K_j$ denote query and key vectors, and $R(\cdot)$ represents an interleaved multi-dimensional rotary encoding $R(p_i)=[R_t(t_i),R_h(h_i),R_w(w_i)]$. This design enables decoupled modeling of temporal evolution and spatial relations while preserving long-range consistency across extended sequences.

Importantly, the forward inference structure remains unchanged. Our contribution lies not in modifying architectural components, but in redefining how structured cases are absorbed during training.

\subsection{Case-Space Modeling Framework}

To establish the empirical support of the case space, we formalize heterogeneous traffic data into a unified structured representation. Each case is defined as a tuple $\mathcal{C} = (x, q, a, e)$, where $x$ denotes the visual context (video clip or image), $q$ represents the natural language query, $a$ is the ground-truth answer, and $e$ signifies an optional explanation. We construct a comprehensive case base by amalgamating multi-source datasets to cover the full spectrum of traffic reasoning. Specifically, the dynamic case repository integrates TrafficQA with a self-collected video subset, totaling approximately 12,000 real-world videos and 70,000 QA pairs. While TrafficQA emphasizes temporal causal reasoning, our self-collected data focuses on abnormal-event identification and semantic consistency via a verification-based format. Complementing this, the static case repository leverages DriveQA, incorporating 448,000 tuples from both real-world traffic signs and CARLA-simulated environments. This diversity, spanning extreme weather conditions (e.g., rain, snow, glare) and varied ego-viewpoints, ensures that the case space encompasses both fine-grained environmental semantics and complex temporal interactions.

Departing from classical CBR systems that rely on explicit online retrieval, our framework utilizes the case base as a unified training substrate to define the empirical distribution over multimodal traffic scenarios. In this paradigm, the case base is not a lookup table for inference but rather a structured manifold for representation learning. By optimizing continuous case embeddings during the training phase, Traffic-MLLM internalizes cross-case structural regularities directly within the model parameters. This enables the model to capture the underlying relationships between heterogeneous cases, facilitating robust adaptation to long-tail scenarios and distribution shifts without the computational overhead of explicit case matching during inference.

\subsection{Curiosity-Driven Case-Space Optimization}

\begin{figure}[!tbp] 
    \centering 
    \includegraphics[width=\linewidth]{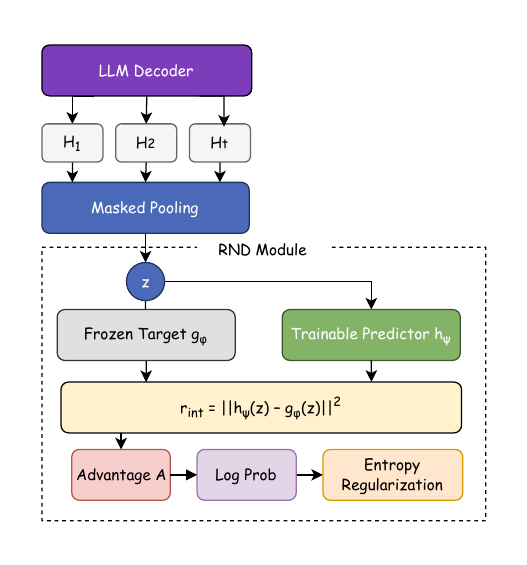} 
    \caption{Curiosity-driven case-space optimization. Latent case representations obtained via masked pooling are evaluated by an RND module to estimate structural novelty. The resulting intrinsic signal adaptively reweights supervision, encouraging the model to allocate learning capacity toward under-represented or uncertain cases.} 
    \label{fig:curiosity} 
\end{figure}

From a case-based reasoning perspective, robust generalization requires high sensitivity to epistemic boundaries within the case manifold. Standard supervised fine-tuning (SFT) typically optimizes likelihood uniformly over the training distribution, which tends to bias learning toward high-frequency patterns while underfitting structurally sparse or boundary cases. Specifically, under the SFT framework, model parameters $\theta$ are optimized by minimizing the negative log-likelihood:
\begin{equation}
\mathcal{L}_{\text{SFT}}(\theta) = -\sum_{t=1}^{T} m_t \log p_\theta(y_t^* \mid y_{<t}^*, x),
\end{equation}
where $m_t$ denotes valid token positions. To address this distribution bias, we formalize each training instance as a structured case within a learned representation space. Let $H=\{H_t\}_{t=1}^{T}$ denote the decoder hidden states; we derive a sequence-level case embedding $z$ through masked pooling:
\begin{equation}
z = \frac{\sum_t m_t H_t}{\sum_t m_t+\epsilon}.
\end{equation}
This embedding $z$ serves as a latent coordinate, effectively locating the current instance within the manifold for subsequent novelty assessment.

To quantify the structural novelty within this latent space, we employ Random Network Distillation (RND), utilizing a frozen, randomly initialized target network $g_\phi$ and a trainable predictor network $h_\psi$. The intrinsic novelty signal $r_{\text{int}}$ is defined by the distillation error:
\begin{equation}
r_{\text{int}} = \|h_\psi(z)-g_\phi(z)\|_2^2.
\end{equation}
To ensure the predictor accurately reflects the current state of the case manifold, it is optimized to minimize the prediction error:
\begin{equation}
\mathcal{L}_{\text{pred}} = \mathbb{E}[r_{\text{int}}].
\end{equation}
In this setup, higher values of $r_{\text{int}}$ highlight cases that are weakly represented or insufficiently modeled, providing a diagnostic signal to focus the model's attention on epistemic gaps.

Rather than introducing the complexity of explicit reinforcement learning, we incorporate this novelty signal as an adaptive reweighting factor. For a greedily generated sequence $a$, we define a clipped novelty advantage $A$ as:
\begin{equation}
A = \operatorname{clip}(r_{\text{int}}-b,-c,c),
\end{equation}
where $b$ is an exponential moving average baseline used to stabilize the scale of the intrinsic reward. The resulting novelty-aware objective is then formulated as:
\begin{equation}
\mathcal{L}_{\text{nov}} = -\mathbb{E}\big[A \log \pi_\theta(a\mid x)\big].
\end{equation}
To preserve representational diversity and prevent mode collapse, we further introduce an entropy regularization term $\mathcal{H}(\pi_\theta)$. The complete optimization objective is thus:
\begin{equation}
\mathcal{L}_{\text{total}} = \mathcal{L}_{\text{SFT}} + \lambda_{\text{nov}}\mathcal{L}_{\text{nov}} + \lambda_{\text{pred}}\mathcal{L}_{\text{pred}} - \lambda_{\text{ent}}\mathcal{H}(\pi_\theta).
\end{equation}
This joint optimization reshapes the internal case manifold by allocating greater modeling capacity to boundary cases, thereby improving robustness under distributional shifts without adding inference-time overhead.

\section{Experiments}
\label{sec:experiments}
\begin{table}[t]
    \centering
    \caption{Accuracy (\%) on \textbf{SUTD-TrafficQA} across different reasoning tasks. Best results are in bold.}
    \label{tab:trafficqa_results}
    \footnotesize
    \setlength{\tabcolsep}{3pt} 
    \renewcommand{\arraystretch}{1.10}

    \resizebox{\linewidth}{!}{%
    \begin{tabular}{lccccccc}
    \toprule
    \textbf{Method} & \textbf{Basic} & \textbf{Attr.} & \textbf{Intro.} & \textbf{Counter.} & \textbf{Fore.} & \textbf{Reverse} & \textbf{All} \\
    \midrule
    Random                        & 25.00 & 25.00 & 25.00 & 25.00 & 25.00 & 25.00 & 25.00 \\
    Eclipse\cite{Xu2021TrafficQA}  & --    & --    & --    & --    & --    & --    & 37.05 \\
    HCRN\cite{HCRN}               & 34.20 & 50.30 & 33.40 & 40.70 & 44.60 & 50.10 & 36.30 \\
    CMCIR\cite{Lin_TPAMI}         & 36.10 & 52.60 & 38.40 & 46.00 & 48.80 & 52.20 & 38.58 \\
    Tem-Adaptor\cite{Tem-adapter} & 46.00 & 47.70 & 34.50 & 55.00 & 36.50 & 44.60 & 46.10 \\
    \midrule
    Qwen2.5-VL\cite{qwen25vl} & 46.50 & 41.60 & 19.70 & 39.20 & 39.10 & 43.90 & 44.10 \\
    Qwen3-VL\cite{qwen3vl} & 48.30 & 44.80 & 22.60 & 42.50 & 41.90 & 46.20 & 46.00 \\
    VideoLLaMA2\cite{videollama2}  & 49.20 & 45.20 & 35.80 & 53.50 & 39.00 & 48.50 & 47.51 \\
    VideoChat2\cite{videochat2}    & 42.50 & 42.80 & 30.40 & 49.20 & 38.10 & 43.80 & 42.17 \\
    Video-LLaVA\cite{videollava}   & 39.70 & 36.40 & 31.10 & 40.50 & 37.20 & 35.80 & 38.35 \\
    DriveMM\cite{drivemm}          & 47.60 & 37.70 & 38.50 & 43.20 & 38.60 & 40.10 & 43.90 \\
    \midrule
    \rowcolor{gray!12}
    \textbf{Traffic-MLLM} & \textbf{48.70} & \textbf{53.20} & \textbf{41.10} & \textbf{57.40} & \textbf{50.20} & \textbf{54.80} & \textbf{50.80} \\
    \bottomrule
    \end{tabular}}
\end{table}

\begin{table*}[t]
    \centering
    \caption{Accuracy (\%) on \textbf{DriveQA-V (CARLA Signs)} across four sign categories. Best results are in bold.}
    \label{tab:driveqa_signs}
    \renewcommand{\arraystretch}{1.20}
    \setlength{\tabcolsep}{5pt}
    \footnotesize
    \begin{tabular}{
        l
        >{\centering\arraybackslash}m{0.9cm}
        >{\centering\arraybackslash}m{1.5cm}
        >{\centering\arraybackslash}m{1.4cm}
        >{\centering\arraybackslash}m{1.2cm}
        >{\centering\arraybackslash}m{2.0cm}}
    \toprule
    \textbf{Model} & \textbf{Size} & \textbf{Regulatory} & \textbf{Warning} & \textbf{Guide} & \textbf{Temporary Control} \\
    \midrule
    Mini-InternVL\cite{InterVL-mini}   & 2B & 64.06 & 55.34 & 65.82 & 45.04 \\
    LLaVA-1.5\cite{LLaVA-1.5}          & 7B & 23.51 & 26.61 & 22.31 & 21.10 \\
    LLaVA-1.6-mistral\cite{LLaVA-1.6}  & 7B & 42.58 & 43.01 & 52.75 & 37.50 \\
    VILA-1.5\cite{VILA}                & 8B & 25.32 & 23.33 & 27.78 & 21.46 \\
    \midrule
    \rowcolor{gray!12}
    \textbf{Traffic-MLLM}         & \textbf{4B} & \textbf{75.65} & \textbf{74.83} & \textbf{72.10} & \textbf{70.58} \\
    \bottomrule
    \end{tabular}
\end{table*}

\subsection{Experimental Setup}
\subsubsection{Datasets}

We evaluate \textit{Traffic-MLLM} on two widely used traffic reasoning benchmarks: TrafficQA and DriveQA, which respectively focus on dynamic video reasoning and static traffic scene understanding.

\textbf{TrafficQA} \cite{Xu2021TrafficQA} is a video-based question–answering benchmark designed for traffic scene understanding. It includes multiple reasoning categories such as event prediction, counterfactual reasoning, attribution analysis, and inverse reasoning. In our experiments, TrafficQA is used to evaluate the model’s capability in spatiotemporal reasoning and causal understanding in dynamic traffic scenarios.

\textbf{DriveQA} \cite{li2024driveqa} is an image-based traffic QA benchmark focusing on traffic sign understanding under diverse environmental conditions. The dataset combines synthetic scenes generated in CARLA with real-world images collected from Mapillary. We use this benchmark to evaluate visual reasoning and cross-domain generalization in traffic environments.

\subsubsection{Baseline Selection}

We compare Traffic-MLLM with representative methods from three categories:
\begin{itemize}
\item classical video question-answering models
\item temporal reasoning architectures
\item recent multimodal large language models (MLLMs)
\end{itemize}

For SUTD-TrafficQA, we include several established traffic video reasoning approaches representing different reasoning paradigms such as hierarchical relational reasoning, cross-modal causal modeling, and temporal adaptation. To provide stronger comparisons with modern vision-language systems, we additionally evaluate recent video-oriented MLLMs capable of performing multimodal reasoning without task-specific architectural modifications. For DriveQA-V and Mapillary evaluation, we further adopt widely used open-source MLLMs with different parameter scales as baselines. This setup enables us to compare Traffic-MLLM against both specialized traffic reasoning architectures and general-purpose multimodal models, while also providing a balanced analysis of model capacity and reasoning performance.

\subsubsection{Implementation Details}

Traffic-MLLM is built upon the Qwen3-VL-4B-Instruct backbone, resulting in approximately 4.2B parameters including multimodal projection layers and the lightweight curiosity-driven RND module. For video reasoning, frames are sampled at 2 FPS with up to 4 frames per clip, and the maximum sequence length is set to 1536 tokens. The RND module consists of a frozen target MLP and a trainable predictor MLP with projection dimension 256, where the intrinsic novelty signal is defined as the squared Euclidean distance between the predictor and target embeddings.

Training is conducted on 4 NVIDIA RTX 3090 GPUs (24GB) using PyTorch distributed training with bf16 mixed precision and gradient checkpointing. We use the AdamW optimizer with a learning rate of $1\times10^{-6}$, per-device batch size 1, gradient accumulation 8 (effective batch size 32), cosine learning rate schedule with 3\% warmup, and zero weight decay. Curiosity-related coefficients are set to: policy weight 0.03, predictor weight 0.3, entropy coefficient $5\times10^{-4}$, advantage clipping threshold 2.0, reward EMA decay 0.99, and consistency weight 0.005. Models are trained for one epoch with early stopping (patience=3), and the best checkpoint is selected based on validation loss. During inference, evaluation is performed with batch size 1 on a single RTX 3090 GPU, with an average latency of approximately \textbf{1500 ms} per sample (0.67 videos/s).

\subsection{Experimental Results}

\begin{table}[t]
    \centering
    \caption{Ablation results on three datasets (Accuracy \%).}
    \label{tab:ablation_all}
    \renewcommand{\arraystretch}{1.15}
    \setlength{\tabcolsep}{6pt}
    \footnotesize
    \begin{tabular}{lccc}
    \toprule
    \textbf{Method} & \textbf{DriveQA-V} & \textbf{Mapillary} & \textbf{TrafficQA} \\
    \midrule
    Baseline (Qwen3-VL) & 68.96 & 74.28 & 46.90 \\
    + Case-based SFT & 72.65 & 80.12 & 49.20 \\
    + Novelty Reweighting (RND) & 73.84 & 81.57 & 50.05 \\
    + Entropy Regularization & 74.31 & 82.42 & 50.42 \\
    \textbf{Full Model} & \textbf{74.77} & \textbf{83.10} & \textbf{50.80} \\
    \bottomrule
    \end{tabular}
\end{table}

\begin{table}[t]
    \centering
    \caption{Ablation results on three datasets (Accuracy \%).}
    \label{tab:ablation_all}
    \renewcommand{\arraystretch}{1.15}
    \setlength{\tabcolsep}{6pt}
    \footnotesize
    \begin{tabular}{lccc}
    \toprule
    \textbf{Method} & \textbf{DriveQA-V} & \textbf{Mapillary} & \textbf{TrafficQA} \\
    \midrule
    Baseline (Qwen3-VL) & 68.96 & 74.28 & 46.90 \\
    + Case-based SFT & 72.65 & 80.12 & 49.20 \\
    + Novelty Reweighting (RND) & 73.84 & 81.57 & 50.05 \\
    + Entropy Regularization & 74.31 & 82.42 & 50.42 \\
    \textbf{Full Model} & \textbf{74.77} & \textbf{83.10} & \textbf{50.80} \\
    \bottomrule
    \end{tabular}
\end{table}

Overall, Traffic-MLLM demonstrates consistent improvements across multiple traffic reasoning benchmarks, outperforming both specialized traffic reasoning models and recent multimodal large language models.

Traffic-MLLM (4B) achieves an overall accuracy of 50.80\% on SUTD-TrafficQA, outperforming prior traffic reasoning architectures such as Tem-Adaptor (46.10\%) by a clear margin. Performance gains are consistently observed across multiple reasoning categories, with particularly strong improvements in counterfactual and inverse reasoning, suggesting that the proposed framework better captures structured interactions among traffic participants. Compared with recent multimodal large language models, including Qwen2.5-VL (44.10\%), Qwen3-VL (46.00\%), VideoLLaMA2 (47.51\%), VideoChat2 (42.17\%), Video-LLaVA (38.35\%), and DriveMM (43.90\%), Traffic-MLLM also achieves the best overall accuracy. These results indicate that simply scaling generic multimodal models is insufficient for complex traffic reasoning tasks, while the proposed case-space learning framework enables more effective structural adaptation to traffic scenarios. Notably, these improvements are achieved with a relatively compact 4B-parameter model, highlighting the efficiency of the proposed curiosity-driven case-space optimization.

On DriveQA-V, Traffic-MLLM achieves the best performance across all four traffic-sign categories despite having fewer parameters than several 7B and 8B baselines, demonstrating the effectiveness of structured case-space learning for domain-specific traffic understanding. Furthermore, on the real-world Mapillary dataset, Traffic-MLLM reaches 78.64\% accuracy in the off-the-shelf setting and further improves to 83.10\% after DriveQA fine-tuning, demonstrating strong cross-domain generalization from synthetic CARLA environments to real-world traffic scenes.

\subsection{Ablation Studies}

We perform ablation studies to evaluate the contribution of each component in the proposed framework. Starting from the same Qwen3-VL backbone, we progressively introduce the structured case-based training strategy and the curiosity-driven case-space optimization. As shown in Table~\ref{tab:ablation_all}, the baseline model achieves 68.96 on DriveQA-V, 74.28 on Mapillary, and 46.90 on TrafficQA. Introducing \textbf{case-based supervised fine-tuning (SFT)} on the multi-source traffic case base significantly improves performance across all datasets, increasing the accuracy to 72.65, 80.12, and 49.20 respectively. This result indicates that organizing heterogeneous traffic samples as structured cases helps the model better align visual observations with traffic semantics and improves domain adaptation.

We then enable the proposed \textbf{curiosity-driven optimization}, where case representations extracted from decoder hidden states are evaluated by an RND module to estimate structural novelty and reweight supervision. Adding \textbf{novelty reweighting} further improves the performance to 73.84 on DriveQA-V, 81.57 on Mapillary, and 50.05 on TrafficQA, suggesting that novelty-aware training encourages the model to allocate more capacity to under-represented or structurally uncertain traffic cases. We additionally introduce \textbf{entropy regularization} to stabilize the reward-weighted optimization and maintain prediction diversity, which provides further improvements across the three benchmarks. Finally, combining all components yields the \textbf{full model}, achieving the best results of 74.77 on DriveQA-V, 83.10 on Mapillary, and 50.80 on TrafficQA. These results confirm that the proposed curiosity-driven case-space optimization effectively complements case-based supervised learning and leads to more robust multimodal traffic reasoning, particularly under cross-domain and long-tail traffic scenarios.

\subsection{Qualitative Analysis}

\begin{figure*}[!tbp] 
    \centering
    \includegraphics[width=0.8\linewidth]{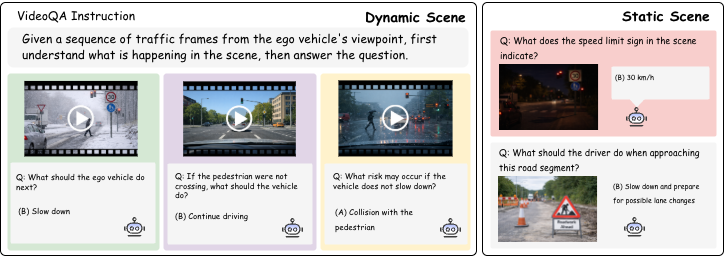}
    \caption{Representative qualitative examples of Traffic-MLLM. The model handles both dynamic traffic reasoning (future actions, counterfactual analysis, and risk prediction) and static scene understanding (traffic sign interpretation), demonstrating consistent reasoning behavior across diverse traffic scenarios.}
    \label{fig:qualitative}
\end{figure*}

Beyond quantitative accuracy, we further analyze the reasoning behavior of \textit{Traffic-MLLM} through qualitative examples. 
Figure~\ref{fig:qualitative} illustrates representative cases covering both dynamic traffic reasoning and static scene understanding, providing insights into how the model forms decisions under different traffic conditions.

In dynamic traffic scenarios, Traffic-MLLM exhibits more stable and consistent reasoning behavior compared with baseline multimodal large language models. 
When facing multi-agent interactions such as vehicle yielding, pedestrian crossing, or potential collision risks, the model tends to maintain coherent reasoning trajectories rather than relying solely on isolated visual cues or dominant temporal patterns. 
This behavior suggests that Traffic-MLLM gradually organizes traffic experiences into an implicit structured case representation during training, allowing the model to reason based on relational patterns among traffic participants. 
When encountering a new traffic situation, the model can map it to structurally similar interaction patterns learned from previous cases, enabling more reliable predictions. 
Moreover, under visually challenging or low-frequency long-tail conditions, Traffic-MLLM still produces relatively stable predictions, indicating that such structural case abstraction improves robustness near distributional boundaries where supervision is sparse.

A similar pattern can be observed in static traffic scene understanding tasks. 
In particular, when transferring from CARLA-generated synthetic training data to real-world Mapillary imagery, Traffic-MLLM maintains more consistent interpretations of traffic signs and regulations, whereas baseline models often exhibit noticeable synthetic bias, such as relying excessively on texture or rendering style. 
In contrast, Traffic-MLLM tends to infer the semantic meaning of traffic signs based on their regulatory roles and contextual relationships within the scene. 
This indicates that the learned case representation captures higher-level semantic structures beyond superficial visual patterns, allowing the model to generalize across visual domains by associating new observations with structurally similar cases in the learned representation space.

Overall, these observations suggest that learning a structured case representation space encourages the abstraction of traffic rules and interaction dynamics, leading to more consistent reasoning, improved robustness in long-tail scenarios, and stronger cross-domain generalization.

\section{Conclusion and Future Work}

In this paper, we presented \textit{Traffic-MLLM}, a curiosity-regularized multimodal framework for traffic scenario reasoning from a case-based perspective. Instead of relying on explicit case retrieval at inference time, the proposed framework learns a structured case representation space directly through supervised training on a multi-source traffic case base. By integrating dynamic traffic videos and static visual question-answering data, the model learns to organize heterogeneous traffic experiences into a unified representation manifold. Furthermore, we introduce a curiosity-driven optimization mechanism based on Random Network Distillation (RND) to emphasize structurally novel or under-represented cases during training. Experimental results on SUTD-TrafficQA and DriveQA demonstrate consistent improvements in dynamic reasoning, regulatory understanding, and cross-domain generalization, showing that representation-level case-space learning provides an effective alternative to explicit retrieval-based CBR for multimodal traffic reasoning.

In future work, we plan to further expand the scale and diversity of the traffic case base by incorporating larger collections of real-world driving videos, simulation data, and long-tail safety-critical scenarios. Increasing the coverage of multimodal traffic experiences may enable the model to better capture rare interaction patterns and improve robustness under distributional shifts. In addition, we aim to extend the current case-space learning paradigm toward world-model-style representations for autonomous driving. By integrating structured case learning with predictive modeling of future states and environment dynamics, we hope to move beyond question-answering tasks and toward more general reasoning and planning capabilities in complex traffic environments.

\bibliographystyle{IEEEtran}
\bibliography{bibliography}

\end{document}